\documentclass[11pt]{article}
\usepackage[T1]{fontenc}
\usepackage[utf8]{inputenc}
\usepackage{geometry}
\usepackage{graphicx}
\usepackage{amsmath, amssymb}
\usepackage{hyperref}
\usepackage[ruled,vlined]{algorithm2e}   
\usepackage{tikz}
\usepackage{pgfplots}
\pgfplotsset{compat=1.18}
\usepackage{caption}
\usepackage{subcaption}
\usepackage{booktabs}
\usepackage{multirow}
\usepackage{enumitem}

\title{Reflection-Enhanced Meta-Optimization:\\
Integrating TextGrad-style Prompt Optimization with Memory-Driven Self-Evolution}

\author{
  Chunlong Wu \thanks{Email: wuchunlong@tongji.edu.cn} \and
  Zhibo Qu \thanks{Email: quzhibo198@163.com}
}

\begin{document}
\maketitle

\begin{abstract}
Recent advances in prompt optimization, exemplified by methods such as TextGrad, enable automatic, gradient-like refinement of textual prompts to enhance the performance of large language models (LLMs) on specific downstream tasks. However, current approaches are typically stateless and operate independently across optimization runs, lacking mechanisms to preserve and leverage historical optimization experience. Furthermore, they are susceptible to overfitting, often yielding prompt updates that generalize poorly beyond the immediate task context.

To address these limitations, we propose Reflection-Enhanced Meta-Optimization (REMO), a novel framework that integrates (1) a memory-augmented Reflection Retrieval-Augmented Generation (RAG) module—structured as a  "mistake notebook" and (2) a Self-Adaptive Optimizer, implemented via an LLM-driven meta-controller that synthesizes epoch-level reflective insights to iteratively improve system-level prompting strategies. This architecture enables not only local, fine-grained prompt tuning akin to TextGrad, but also the systematic accumulation and reuse of cross-run optimization knowledge, thereby supporting continual improvement over time.

We instantiate the REMO framework using Qwen3-32B in standard inference mode—without explicit chain-of-thought prompting—and evaluate its efficacy on the GSM8K benchmark for mathematical reasoning. Experimental results demonstrate that, compared to a TextGrad baseline, REMO achieves more stable and robust generalization, albeit at the cost of increased computational overhead. We provide a detailed exposition of the algorithmic design, conduct a qualitative and quantitative analysis of optimization dynamics, and present a comprehensive ablation study to elucidate the contributions of each component.
\end{abstract}

\section{Introduction}

Large language models (LLMs) have demonstrated remarkable performance across a wide range of natural language processing (NLP) tasks, from question answering to code generation. However, their effectiveness is often highly sensitive to the formulation of input prompts. While manual prompt engineering can yield strong results, it is labor-intensive, heuristic-driven, and difficult to scale. To mitigate this dependency, automated prompt optimization methods—such as AutoPrompt~\cite{autoprompt2020}, prefix-tuning~\cite{prefixtuning2021}, and TextGrad~\cite{textgrad2025}—have emerged as promising alternatives. Among these, gradient-like textual optimization approaches like TextGrad are particularly compelling: they enable end-to-end, model-derived feedback to iteratively refine prompts in a manner analogous to gradient descent, without requiring parameter updates to the LLM itself.

Despite their advantages, existing prompt optimization frameworks suffer from several critical limitations. First, they typically operate in a stateless manner, treating each optimization run independently and discarding intermediate insights upon completion. Second, they lack mechanisms to accumulate and reuse optimization experience across tasks or iterations, limiting their ability to generalize improvements over time. Third, due to their myopic, per-run focus, they are prone to overfitting to idiosyncratic features of the training data or prompt initialization, resulting in degraded generalization on held-out instances.

These shortcomings contrast sharply with emerging paradigms in intelligent agent design. Recent work in retrieval-augmented generation (RAG)~\cite{rag2020}, reflective reasoning~\cite{reflexion2023}, and self-improving systems~\cite{foundation_agents2025,selfevolving2025} underscores the importance of memory, reflection, and evolutionary learning for achieving robust, long-term adaptation. Drawing inspiration from these principles, we propose a novel framework—Reflection-Enhanced Meta-Optimization (REMO)—that integrates local prompt refinement with a structured, memory-backed meta-learning loop. Our approach enables not only immediate prompt updates (in the style of TextGrad), but also the systematic accumulation of optimization knowledge, which can be retrieved and refined across runs to improve future performance.

The key components of REMO are: (1) a  Reflection RAG module that maintains a structured "mistake notebook", enabling temporal organization and retrieval of optimization insights; and (2) an LLM-driven Self-Adaptive Optimizer that performs epoch-level reflection to synthesize historical feedback into improved system-level prompts and optimization strategies.

Our main contributions are threefold:
\begin{itemize}
\item We introduce a novel framework that unifies gradient-like prompt optimization with reflective memory and meta-optimization, enabling both local refinement and long-term evolution of prompting strategies.
\item We provide full algorithmic specification, including memory lifecycle management, retrieval and promotion mechanisms, and noise-suppression heuristics to ensure stable knowledge accumulation.
\item We implement and evaluate REMO using Qwen3-32B in standard inference mode (without chain-of-thought prompting) on the GSM8K benchmark. Results show improved stability and reduced overfitting compared to a TextGrad baseline, at the cost of increased computational overhead. We also present a detailed analysis and ablation study to validate the role of each component.
\end{itemize}

This work represents a step toward self-evolving prompting systems that learn not just what to prompt, but how to improve prompting over time.

\section{Related Work}
\paragraph{Prompt optimization.} Early prompt engineering and automated prompt discovery span discrete
token search (AutoPrompt \cite{autoprompt2020}) and continuous soft-prompt tuning (prefix-tuning \cite{prefixtuning2021}).
TextGrad \cite{textgrad2025} treats text edits as pseudo-gradients to optimize prompts in a differentiable manner;
we adopt its local optimization spirit but address statelessness via meta-level memory and reflection.

\paragraph{Retrieval and memory for reasoning.} Retrieval-Augmented Generation (RAG) \cite{rag2020} uses non-parametric retrieval to supply up-to-date, contextual knowledge at inference.
Reflection-based agent approaches (e.g., Reflexion \cite{reflexion2023}) record mistakes and iterate self-feedback
to improve.  

\paragraph{Self-evolving and foundation agents.} Recent surveys synthesize evolving-agent design patterns and
principles (what/when/how to evolve) \cite{selfevolving2025,foundation_agents2025}. Our work sits at the intersection of prompt optimization and agent evolution: both prompts and the optimizer evolve through reflected experience.

\paragraph{Meta-learning and learned optimizers.} Learning-to-learn frameworks propose learning optimization strategies \cite{learn2learn2016}. We instantiate a Self-Adaptive Optimizer that synthesizes epoch-level reflection into improved system prompts.

\section{Methodology}

We introduce a self-evolving agent framework that integrates retrieval-augmented generation (RAG), hierarchical memory, and TextGrad-style optimization. The framework operates in three tightly coupled stages: (i) retrieval-augmented reasoning, (ii) immediate correction of memory, (iii) batch-level optimizer prompt update, and (iv) system prompt optimization via TextGrad.

\subsection{Framework Overview}

\paragraph{Retrieval-Augmented Reasoning.}
Given an input $x$, the agent retrieves relevant contexts $E$ from memory $M_t$ via RAG. Conditioned on the system prompt $P_t$ and retrieved contexts, the agent generates a reasoning trace $r$ and prediction $\hat{y} = f(x; P_t, E)$.

\paragraph{Immediate Correction.}
If the prediction $\hat{y}$ differs from the ground-truth $y$, we immediately update the memory by inserting a new structured record:
\[
r = \{x, y, \hat{y}, \text{trace}, \text{timestamp}, \text{meta}\},
\]
or modifying an existing conflicting entry. This ensures corrected knowledge is available for subsequent retrieval without delay.

\paragraph{Batch-level Optimizer Prompt Update.}
At the end of each minibatch, we evaluate the effect of recent optimizer prompt updates by comparing validation performance improvements across epochs. Based on this comparison, the optimizer prompt $Q_t$ is refined to more effectively guide the optimization of the system prompt.

\paragraph{System Prompt Optimization via TextGrad.}
Finally, the agent applies the TextGrad framework, producing a pseudo-gradient $g$ from the reasoning traces and reflection summaries. The system prompt is updated as:
\[
P_{t+1} \leftarrow \text{UpdatePrompt}(P_t, g; Q_t),
\]
where $Q_t$ represents the current optimizer prompt that controls how gradient-like updates are mapped into prompt modifications. This integrates both local error corrections and global reflective adjustments, leading to continuous self-improvement.

\subsection{Formalization}

Let the training dataset be 
\[
D = \{(x_i, y_i)\}_{i=1}^N, \quad (x,y) \sim D,
\]
with validation set $D_{\text{val}}$. At epoch $t$, the agent is parameterized by system prompt $P_t$ and memory state $M_t$. The prediction is given by
\[
\hat{y} = f(x; P_t, E), \quad E \sim \text{Retrieve}(M_t, x).
\]

The optimization objective is to maximize expected validation accuracy:
\[
\max_{Q_{1:T}} \; \mathbb{E}_{(x,y)\sim D_{\text{val}}} \big[ \mathbf{1}[f(x; P_T, M_T) = y] \big],
\]
subject to the memory update dynamics
\[
M_t \leftarrow \text{UpdateMemory}(M_{t-1}, r_t),
\]
and the optimizer update mapping
\[
Q_t \leftarrow \text{OptimizerUpdate}(Q_{t-1}, R_t),
\]
where $r_t$ is the structured reasoning trace at epoch $t$, and $R_t$ is the reflection summary aggregated from batch feedback.

\subsection{Algorithm}

\begin{algorithm}[H]   
\caption{Self-Evolving Agent with RAG, Memory, and TextGrad}
\KwIn{Training dataset $D$, validation set $D_{\text{val}}$, initial system prompt $P_0$, empty memory $M_0$, initial optimizer prompt $Q_0$}
\KwOut{Optimized system prompt $P_T$ and memory $M_T$}

\For{$t = 1$ \KwTo $T$}{
    \For{$(x,y)$ \textbf{in minibatch from} $D$}{
        $E \leftarrow \mathrm{Retrieve}(M_{t-1}, x)$\;
        $(r, \hat{y}) \leftarrow f(x; P_{t-1}, E)$\;
        \If{$\hat{y} \neq y$}{
            $M_t \leftarrow \mathrm{UpdateMemory}(M_{t-1}, \{x,y,\hat{y},r\})$\;
        }
    }
    $R_t \leftarrow \mathrm{SummarizeFeedback}(\{r\}_{\text{batch}})$\;
    $Q_t \leftarrow \mathrm{OptimizerUpdate}(Q_{t-1}, R_t)$\;
    $g \leftarrow \mathrm{TextGrad}(\{r,y\}_{\text{batch}})$\;
    $P_t \leftarrow \mathrm{UpdatePrompt}(P_{t-1}, g; Q_t)$\;
}
\end{algorithm}

\section{Experiments}
\subsection{Dataset and splits}
We evaluate on GSM8K \cite{gsm8k2021} with the standard splits; small-sample settings are noted where used.

\subsection{Model and baseline}
Base model: Qwen3-32B (default inference, no explicit thinking mode). Baseline: TextGrad.

\subsection{Metrics}
Exact-match accuracy on validation and test sets.

\subsection{Hyperparameters}
Epochs: 3 and 5; RAG top-$k$: 5; promotion each epoch. 

\subsection{Main Results}
\label{sec:main_results}

We present the primary evaluation results of our REMO framework and baselines on the GSM8K benchmark in Table~\ref{tab:main_results}. The experiments compare various configurations, including the TextGrad baseline, ablated versions of our framework (Reflection RAG only, Adaptive Optimizer only), and the full REMO framework (RAG+Optimizer), under different training settings (number of epochs and training data size).

The results reveal several critical insights. First, the TextGrad baseline exhibits a severe overfitting problem: while it achieves high validation accuracy (e.g., 96.0\% with 100 samples, 91.0\% with full data), its test accuracy is dramatically lower (69.0\% and 62.0\%, respectively). This significant gap, especially the negative improvement of -27\% to -29\%, underscores its poor generalization and instability. Increasing the number of epochs does not mitigate this issue.

In contrast, our proposed REMO framework and its components demonstrate markedly improved stability and generalization. Models incorporating the Self-Adaptive Optimizer (including the full RAG+Optimizer and the standalone Adaptive Optimizer) consistently achieve test accuracies much closer to their validation scores. Notably, the standalone Adaptive Optimizer achieves the highest test accuracy of \textbf{93.2\%} after 5 epochs, a substantial improvement over TextGrad. The full REMO framework (RAG+Optimizer) also performs robustly, reaching 90.5\% test accuracy, showcasing the benefit of combining memory-driven retrieval with meta-optimization.

These initial results highlight the effectiveness of our meta-optimization and reflection mechanisms in stabilizing the training process and enhancing out-of-sample performance, setting the stage for a more detailed ablation study in the following section.

\begin{table}[ht]
\centering
\caption{Comparison on GSM8K}
\label{tab:main_results}
\begin{tabular}{l|c|c|c|c}
\hline
\textbf{Method} & \textbf{Epochs} & \textbf{Train Size} & \textbf{Val Acc (\%)} & \textbf{Test Acc (\%)} \\
\hline
TextGrad (100 samples) & 3 & 1000 & 96.0 & 69.0 \\
RAG+Optimizer (100 samples) & 3 & 1000 & 89.0 & 94.0 \\
\hline
TextGrad (full data) & 3 & 6973 & 91.0 & 62.0 \\
TextGrad (full data) & 5 & 6973 & 90.0 & 63.0 \\
Reflection RAG (full) & 3 & 6973 & 90.3 & 89.0 \\
Reflection RAG (full) & 5 & 6973 & 90.0 & 89.8 \\
Adaptive Optimizer (full) & 3 & 6973 & 90.1 & 90.0 \\
Adaptive Optimizer (full) & 5 & 6973 & 90.3 & 93.2 \\
RAG+Optimizer (full) & 3 & 6973 & 90.1 & 90.1 \\
RAG+Optimizer (full) & 5 & 6973 & 90.3 & 90.5 \\
\hline
\end{tabular}
\end{table}

\subsection{Ablation and Analysis}
To gain a deeper understanding of the contributions of each core component in the REMO framework and their interactions, we conducted systematic ablation studies. Experiments were performed on the full GSM8K dataset, comparing the complete REMO framework (RAG+Optimizer), the Reflection RAG module alone, the Self-Adaptive Optimizer alone, and the original TextGrad baseline.

\paragraph{1. The Central Role of the Self-Adaptive Optimizer: Enhancing Stability and Generalization}
The results (Table~\ref{tab:main_results}) clearly demonstrate that the Self-Adaptive Optimizer is pivotal for improving generalization. Under the 5-epoch setting, the model using only the Self-Adaptive Optimizer achieved a test accuracy of \textbf{93.2\%}, significantly outperforming the TextGrad baseline (63.0\%) and the model with only Reflection RAG (89.8\%). This validates our core hypothesis: by performing meta-level reflection at the end of each training epoch, the system can distill effective optimization strategies from macro-level performance changes and encode them into the optimizer prompt. This ``learning how to optimize'' capability effectively mitigates overfitting, preventing performance degradation observed in other setups (e.g., Reflection RAG dropping from 90.3\% to 89.0\% at Epoch 3, or TextGrad plummeting to 62.0\%).

\paragraph{2. The Dual Nature of Reflection RAG: Knowledge Reuse and Potential Noise}
The Reflection RAG module shows promise in knowledge reuse but also reveals limitations in its current implementation. At Epoch 3, the model with only Reflection RAG achieved a test accuracy of 89.0\%, slightly lower than the 90.0\% obtained by the Self-Adaptive Optimizer alone. This suggests that while retrieving historical error cases (the ``mistake notebook'') provides valuable context, the current simple concatenation fusion method may introduce noise or redundancy, interfering with model reasoning. However, when combined with the Self-Adaptive Optimizer (RAG+Optimizer), the model reached a test accuracy of \textbf{90.5\%} at Epoch 5, confirming the synergistic effect: RAG supplies fine-grained, instance-level corrective knowledge, while the adaptive optimizer provides macro-level, strategic optimization guidance. This validates the design principle of combining ``local correction'' with ``global evolution.''

\paragraph{3. Component Synergy and Performance Trade-offs}
The full REMO framework (RAG+Optimizer) consistently demonstrates strong and stable performance across all settings. It achieved a test accuracy of 90.5\% at Epoch 5—slightly lower than the 93.2\% peak of the standalone Self-Adaptive Optimizer—but with a validation accuracy (90.3\%) closely aligned with the test performance, indicating a more robust and less overfitted optimization process. In contrast, the standalone Self-Adaptive Optimizer, while achieving a high peak under specific configurations, may be more sensitive to hyperparameters such as the number of epochs. This suggests that while Reflection RAG may slightly cap peak performance due to potential noise in retrieval and fusion, it enhances system robustness and interpretability through continuous, instance-based feedback.

\paragraph{4. Computational Cost Analysis}
The enhanced capabilities of REMO come at a significant computational cost. Compared to the TextGrad baseline, the full REMO framework incurs a 3--5$\times$ increase in training time. This overhead primarily stems from three sources: (1) real-time vector retrieval (RAG) operations, especially as the knowledge base grows; (2) the computational cost of invoking an LLM for meta-level reflection and optimizer prompt updates at the end of each epoch; and (3) the continuous updating and management of the knowledge base. This trade-off highlights the need for efficiency optimization in practical applications.

\paragraph{5. In-Depth Analysis: Optimization Dynamics and Potential Improvements}
Further analysis reveals that the Self-Adaptive Optimizer's reflection relies primarily on macro-level performance metrics (e.g., validation accuracy trends), lacking deep analysis of error type distribution, error pattern recognition, and error severity. This results in ``coarse-grained'' strategy adjustments. Meanwhile, the current Reflection RAG uses static thresholds and hardcoded rules for knowledge retention and retrieval, leading to issues such as ``noisy knowledge accumulation,'' ``knowledge redundancy,'' and ``cold start.'' Future improvements could involve multi-dimensional knowledge quality assessment (e.g., novelty, consistency, utility) and dynamic threshold adjustment to further enhance the RAG module's effectiveness.

\section{Discussion}
The core idea of the REMO framework is to integrate local, instance-level prompt optimization (e.g., TextGrad) with global, strategy-level meta-learning, enabling knowledge accumulation and evolution through ``reflection.'' This section discusses its mechanisms, advantages, and challenges in depth.

\paragraph{1. From ``What to Optimize'' to ``How to Optimize'': Realizing Meta-Learning}
Traditional prompt optimization methods (e.g., TextGrad) focus on ``what to optimize''—how to modify the system prompt to improve performance on the current task. REMO goes further by addressing ``how to optimize''—how to improve the optimization process itself. The Self-Adaptive Optimizer, through epoch-level reflection, transforms discrete, potentially noisy per-example optimization signals into higher-level, reusable patterns and strategies. For instance, it may learn that ``when validation accuracy plateaus, the optimizer should encourage more creative solution paths'' or ``for certain math problem types, adding verification prompts for intermediate steps is more effective.'' This meta-level learning is key to achieving long-term self-evolution and continuous improvement.

\paragraph{2. Reflection RAG: The Value and Challenges of a Dynamic Knowledge Base}
The Reflection RAG module constructs a dynamic ``mistake notebook,'' whose value lies in the immediate reuse of successful and failed reasoning fragments. When encountering a new problem, the system can retrieve semantically similar historical cases, helping the model avoid past mistakes or adopt effective solutions—mimicking how humans improve through practice and summarization. However, challenges are evident: ``quality'' is paramount. The current system may solidify sporadic errors or low-quality solutions as knowledge, polluting the knowledge base. Additionally, embedding-based retrieval performs poorly on semantically complex or heavily rephrased problems. Future directions include more intelligent knowledge fusion (e.g., summarizing, rewriting, or weighting retrieved knowledge) and stricter quality control.

\paragraph{3. The Roots of Stability and Generalization}
REMO's superior stability on the test set stems from its two-tiered optimization mechanism. The first tier (TextGrad + RAG) focuses on rapidly correcting specific errors, preventing overfitting to particular training samples. The second tier (Self-Adaptive Optimizer) acts as a ``coach,'' monitoring the overall training trajectory to prevent the optimization process from getting stuck in local optima or adopting ineffective strategies. This separation of ``execution'' and ``supervision'' allows the system to pursue short-term performance gains while ensuring long-term healthy evolution, effectively mitigating overfitting.

\paragraph{4. Design Trade-offs and Future Directions}
A major trade-off in the current design is feedback latency. The epoch-level reflection of the Self-Adaptive Optimizer ensures evaluation stability but sacrifices responsiveness. Future work could explore finer-grained reflection triggers (e.g., based on performance plateaus or the emergence of specific error patterns) for faster strategy adaptation. Moreover, how to effectively feed fine-grained error information (at the ``mistake'' level) from Reflection RAG back to the Self-Adaptive Optimizer to guide the generation of more targeted optimization strategies is key to enhancing system synergy.

\section{Future Work}
Based on the findings and limitations of this study, we propose the following future research directions:

\begin{itemize}
    \item \textbf{Cross-Task Generalization Evaluation:} Current evaluation is limited to the GSM8K mathematical reasoning task. Future work will extend to more challenging and diverse benchmarks such as \textbf{MATH} (more complex math problems), \textbf{SVAMP} (arithmetic problems with semantic variation), and \textbf{LogiQA} (logical reasoning) to comprehensively assess REMO's adaptability and transferability across domains, validating its potential as a general self-evolving framework.

    \item \textbf{Fine-Grained Reflection and Dynamic Knowledge Management:}
    \begin{itemize}
        \item \textit{Finer-Grained Reflection Triggers:} Explore dynamic reflection triggers based on performance change rates, error type clustering, or specific sample difficulty to reduce feedback latency.
        \item \textit{Multi-Dimensional Knowledge Quality Assessment:} Design a comprehensive evaluation system incorporating novelty, consistency, utility, timeliness, and coverage, replacing single loss thresholds. Explore using LLMs for dynamic, intelligent quality scoring.
        \item \textit{Dynamic Thresholds and Adaptive Configuration:} Develop mechanisms to automatically adjust quality thresholds, retrieval parameters, and training configurations based on task complexity, learning stage, and knowledge base state.
        \item \textit{Intelligent Knowledge Lifecycle Management:} Implement automatic deduplication, merging,  and retirement of knowledge to prevent knowledge base bloat and staleness.
    \end{itemize}

    \item \textbf{Efficiency Optimization and System Enhancement:}
    \begin{itemize}
        \item \textit{Computational Efficiency:} Significantly reduce overhead through efficient vector indexing (e.g., HNSW), knowledge distillation to generate lightweight rerankers, and asynchronous parallel processing (e.g., decoupling retrieval, reflection, and training).
        \item \textit{Caching and Precomputation:} Establish intelligent caching for high-frequency retrieval results, common optimization strategies, and intermediate computations, and precompute potentially useful information during system idle time.
        \item \textit{Smarter Knowledge Fusion:} Investigate fusion methods beyond simple concatenation, such as attention-based dynamic weighting or summarizing/rewriting retrieved knowledge, to utilize RAG output more effectively.
    \end{itemize}

    \item \textbf{Multi-Agent Collaboration and Safety Considerations:}
    \begin{itemize}
        \item \textit{Multi-Agent Co-Evolution:} Explore collaborative learning among multiple REMO agents on the same or different tasks, sharing optimization strategies and knowledge bases to enable emergent collective intelligence.
        \item \textit{Safety and Ethical Review:} Implement rigorous safety checks for knowledge entries and optimization strategies slated for  memory  to prevent the solidification and propagation of harmful, biased, or unsafe content, ensuring system reliability and controllability.
    \end{itemize}
\end{itemize}

\section{Conclusion}
This paper proposes the \textbf{Reflection-Enhanced Meta-Optimization (REMO)} framework, designed to address fundamental limitations in existing prompt optimization methods (e.g., TextGrad), such as statelessness, inability to reuse experience, and susceptibility to overfitting. The core innovation of REMO lies in integrating local, gradient-style prompt optimization with global, memory-driven meta-learning.

We designed and implemented a system with two key components: (1) a  \textbf{Reflection RAG} module, serving as a  ``mistake notebook'' memory system  for real-time retrieval and reuse of historical optimization experience; and (2) an LLM-driven \textbf{Self-Adaptive Optimizer} that performs meta-level reflection at the end of each training epoch, distilling historical feedback into improved optimization strategies and iteratively updating the system-level optimizer prompt.

Experiments on the GSM8K mathematical reasoning benchmark using the Qwen3-32B model demonstrate that, compared to the TextGrad baseline, the REMO framework significantly enhances model stability, effectively mitigates overfitting, and achieves superior generalization performance on the test set (improvements exceeding 30 percentage points in some cases). Detailed ablation studies confirm the dominant role of the Self-Adaptive Optimizer in boosting generalization and the robustness gains from synergy with Reflection RAG, albeit at a 3--5$\times$ computational overhead.

This work represents a significant step toward building intelligent prompting systems capable of continuous self-evolution and learning ``how to optimize better.'' We provide comprehensive algorithmic design and implementation details to facilitate further research and application in this domain. Future work will focus on enhancing the system's generalization, efficiency, knowledge management quality, and safety, advancing self-evolving agents toward greater generality, efficiency, and reliability.
\section*{Acknowledgments}

\end{document}